\documentclass{article}
\pdfoutput=1

\usepackage[final]{neurips_2022}

\usepackage[utf8]{inputenc} %
\usepackage[T1]{fontenc}    %
\usepackage{hyperref}       %
\usepackage{url}            %
\usepackage{booktabs}       %
\usepackage{amsfonts}       %
\usepackage{nicefrac}       %
\usepackage{microtype}      %
\usepackage{xcolor}         %
\usepackage{wrapfig}
\usepackage[numbers]{natbib}

\usepackage{amsmath,amsfonts,bm}
\usepackage{multirow}
\usepackage{tabularx}
\usepackage{graphicx}
\usepackage{adjustbox}
\usepackage{amsthm}

\newcommand{\mset}[1]{\left\{\kern-.5em\left\{ #1 \right\}\kern-.5em\right\}}
\newcommand{\mmset}[1]{\{\kern-.4em\{ #1 \}\kern-.4em\}}

\newcommand{\norm}[1]{\left\Vert#1\right\Vert}

\newcommand{\abs}[1]{\left\vert#1\right\vert}

\newcommand{\set}[1]{\left\{#1\right\}}
\newcommand{\parr}[1]{\left (#1\right )}
\newcommand{\brac}[1]{\left [#1\right ]}

\newcommand{\Real}{\mathbb R}

\newcommand{\eps}{\varepsilon}
\newcommand{\too}{\rightarrow}

\newcommand{\rebb}[1]{{\color{black}{}#1{}}}

\newcommand{\eg}{{e.g.}}
\newcommand{\ie}{{i.e.}}

\def\eqref#1{equation~\ref{#1}}

\def\1{\bm{1}}

\def\eps{{\epsilon}}

\def\vec1{{\bm{1}}}

\DeclareMathAlphabet{\mathsfit}{\encodingdefault}{\sfdefault}{m}{sl}
\SetMathAlphabet{\mathsfit}{bold}{\encodingdefault}{\sfdefault}{bx}{n}

\def\gC{{\mathcal{C}}}

\def\gG{{\mathcal{G}}}

\def\gL{{\mathcal{L}}}

\def\gS{{\mathcal{S}}}

\title{VisCo Grids: Surface Reconstruction \\with Viscosity and Coarea Grids}

\author{%
  Albert Pumarola$^{*1}$,
  Artsiom Sanakoyeu$^{*1}$,
  Lior Yariv$^{2}$,
  Ali Thabet$^{1}$,
  Yaron Lipman$^{1,2}$ \\
$^{1}$Meta AI, $^{2}$Weizmann Institute of Science\\
}

\begin{document}

\maketitle
\def\thefootnote{*}\footnotetext{These authors contributed equally to this work.}

\begin{abstract}
  Surface reconstruction has been seeing a lot of progress lately by utilizing Implicit Neural Representations (INRs). Despite their success, INRs often introduce hard to control inductive bias (i.e., the solution surface can exhibit unexplainable behaviours), have costly inference, and are slow to train.  The goal of this work is to show that replacing neural networks with simple grid functions, along with two novel geometric priors achieve comparable results to INRs, with instant inference, and improved training times. To that end we introduce VisCo Grids: a grid-based surface reconstruction method incorporating Viscosity and Coarea priors. Intuitively, the Viscosity prior replaces the smoothness inductive bias of INRs, while the Coarea favors a minimal area solution. Experimenting with VisCo Grids on a standard reconstruction baseline provided comparable results to the best performing INRs on this dataset.

\end{abstract}
\section{Introduction}
\begin{wrapfigure}[17]{r}{0.57\textwidth}\vspace{-17pt}
  \begin{center}
    \includegraphics[width=0.52\textwidth]{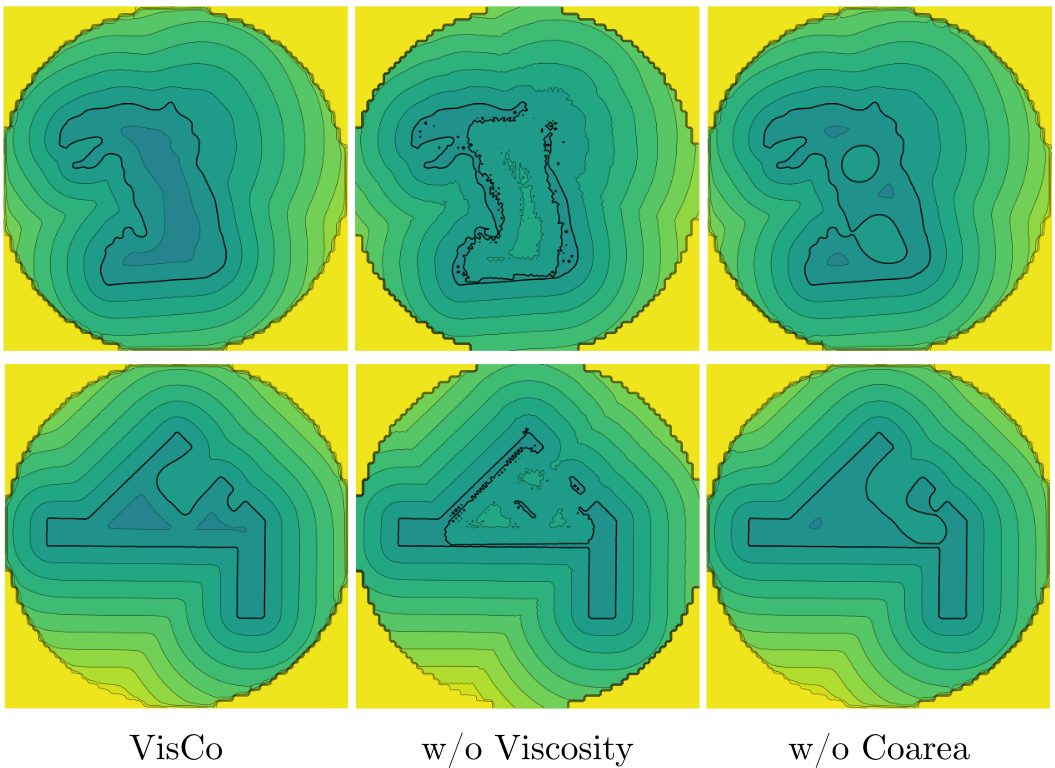}\vspace{-10pt}
  \end{center}
  \caption{The VisCo prior (left) is incorporating viscosity and Coarea; Middle and right shows ablations on each.}\label{fig:teaser}
\end{wrapfigure}
Reconstructing 3D surfaces from sparse point clouds is a long standing problem in both computer vision and graphics \cite{berger2017survey}. Methods tackling this problem aim to estimate 3D surfaces given as input unordered point sets (point clouds), with or without corresponding normals. Surface representations can be divided to two groups: parametric and implicit. Parametric methods represents the surface using some parametric domain, while implicit methods represent the surface as some level-set, $\gS=\set{p\in\Real^3\vert f(p)=c}$, of a volumetric function $f:\Real^3\too \Real$. While parametric methods can easily sample the surface, implicit methods can readily adapt to topological changes of the reconstructed surface. Parametric methods include, \eg, meshes and spline surfaces, while implicit methods use, \eg, volumetric data structures such as voxel grids, Radial Basis Functions (RBFs), or (recently) neural networks. 

Implicit Neural Representation (INR) \cite{mescheder2019occupancy, park2019deepsdf,chen2019learning,atzmon2019sal, erler2020points2surf, gropp2020implicit,atzmon2021sald, peng2021shape} is categorized as an implicit method using a neural networks to define the implicit function $f$. INRs build upon the inherit inductive bias in neural networks and their optimization process to provide smooth yet flexible and expressive surface reconstructions. INRs have several disadvantages: First, the neural inductive bias is hard to control, often introducing undesired or unexplained surface behaviours. In fact, a considerable research effort is dedicated to fix/change/control this bias \cite{tancik2020fourier,sitzmann2020implicit,lipman2021phase,lindell2021bacon}. Second, INRs have an increased deployment cost, requiring many network evaluations for surface contouring, \eg, with marching cube based methods \cite{mescheder2019occupancy, park2019deepsdf}, or direct rendering \cite{yariv2020multiview, mildenhall2020nerf}. Lastly, although using high optimized solvers, INRs are still slow to train.  

The goal of this work it show that network-free grid-based implicit representations can achieve INR-level reconstructions when incorporating suitable priors. To that end, we present VisCo Grids: a grid-based surface reconstruction algorithm that incorporates well-defined geometric priors: Viscosity and Coarea. In short, VisCo, see Figure \ref{fig:teaser}, right. The viscosity loss, is replacing the Eikonal loss \cite{gropp2020implicit,sitzmann2020implicit} used in INRs for optimizing Signed Distance Functions (SDF). The Eikonal loss posses many bad minimal solution that are avoided in the INR setting due to the network's inductive bias, but are present in the grid parametrization, see \eg, Figure \ref{fig:teaser}, middle.  The viscosity loss, uses the notion of vanishing viscosity to regularize the Eikonal loss and provide well defined smooth solution that converges to the "correct" viscosity SDF solution. The viscosity loss provides smooth SDF solution but do not punish excessive or "ghost" surface parts, see \eg, Figure \ref{fig:teaser} (right). Therefore, a second useful prior is the coarea loss, directly controlling the surface's area, and encourages it to be smaller. The coarea loss is defined using a "squashing" function applied to the viscosity SDF making it approximately an indicator function, and then integrates its gradient norm over the domain. Integrating the gradient norm of a function is called the Total Variation loss \citep{Chambolle10anintroduction,lipman2021phase} and is measuring the perimeter of indicator functions, which in our case approximates $\mathrm{area}(\gS)$. VisCo grids (as other grid methods) have instant inference, and even with our current rather naive implementation are faster to train than INRs. Considerable training time improvement are expected with a more efficient implementation. 

We tested VisCo Grids on a standard 3D reconstruction dataset, and achieved comparable accuracy to the state-of-the-art INR methods. Through ablations, we demonstrate the properties and benefit in the VisCo prior. 

\section{Related Work}

\paragraph{3D Surface Reconstruction} 
Classical approaches for surface reconstruction from point clouds \cite{berger2017survey} are either parametric \cite{amenta1998new} or implicit with mostly linear function bases, \eg, grids or radial basis functions \ \cite{carr2001reconstruction,kazhdan2006poisson}. 
Recent works have developed methods for surface reconstruction using neural networks, which consist of a non-linear function space, making these methods non-convex. Those methods differ by how they choose to represent the 3D reconstruction. \cite{groueix2018papier,williams2019deep,hanocka2020point2mesh} employ a parametric point of view. Such discretizations do not yield watertight reconstruction, and/or lack topological detail. A more flexible solution is the Implicit Neural Representation (INR). INRs based methods \cite{park2019deepsdf,mescheder2019occupancy,atzmon2019sal,chen2019learning,gropp2020implicit,sitzmann2020implicit,lipman2021phase,ben2022digs} show great progress in leveraging the inductive bias of MLPs to represent smooth surfaces, using additional losses and regularizers. \rebb{For example, \cite{lipman2021phase} introduce a perturbed Dirichlet loss (\ie, norm of gradient) to push for a unique and regular occupancy solution; \cite{ben2022digs} incorporates a Divergence loss (\ie, absolute value of the divergence of the gradient of trained distance field) for encouraging the learned field to  resemble a gradient field of a true distance functions. } 
Neural Spline \cite{williams2021neural} does not use neural networks directly, rather derive a kernel-based formulation arising from infinitely-wide shallow networks. Shape As points (SAP) \cite{peng2021shape} represent the surface using a differentiable Poisson solver and contouring process.

\paragraph{Grid-based representations} Recent works suggested to reduce, completely or partially, the use of neural networks in implicit representations and replacing it with a grid-based data structure. This is due to the heavy computational resources required in optimizing and evaluating neural networks. Plenoxels \cite{yu2021plenoxels} propose a view-dependent sparse voxel model and show comparable results to NeRF \cite{mildenhall2020nerf} and a speedup of two orders of magnitude. Neural Geometric Level of Detail \cite{takikawa2021nglod} uses an octree-based feature volume and a small MLP to represent SDF.  \cite{mueller2022instant} shows fast training of INR's using a small neural network augmented by a multiresolution hash table with trainable features. Similar to DeepSDF \cite{park2019deepsdf}, both work used 3D supervision for learning the SDF. 
\section{Method}
\label{s:method}

We consider the 3D euclidean space $\Real^3$ with points $p=(x,y,z)\in\Real^3$. We discretize the unit cube $\gC=[0,1]^3$ with a 3D voxel grid $\gG=\set{p_I}$, with nodes $p_I$ indexed by $I=(i,j,k)$, $i,j,k\in [n]=\set{1,\ldots,n}$, \ie, $p_I=(x_{ijk},y_{ijk},z_{ijk})$. We denote by $h=n^{-1}$, and by $N=n^3$ the total number of nodes.   
We represent our reconstructed surface as a zero level of a scalar function $f$ defined over the cube $\gC$. $f$ is defined by prescribing its values at the grid's nodes $f_I\in\Real$ and trilinear interpolating in each voxel. We will denote by $f(p)$ the interpolated value at point $p$. 

Given an input point cloud consisting of $m$ points $q_k\in\Real^3$ with or without (unit norm) normals $n_k\in \Real^3$, $k\in [m]$, our goal is to compute $f$ so that its zero level set approximates the unknown surface, \ie, 
\begin{equation}
    \gS = \set{p\in\gC \ \vert \ f(p)=0}.
\end{equation}
Our approach to compute $f$ is to minimize a loss function of the form
\begin{equation}
    \gL = \gL_{\text{data}} + \gL_{\text{prior}}
\end{equation}
where 
\begin{equation}\label{e:loss_data}
    \gL_{\text{data}} = \frac{\lambda_{\text{p}}}{m}\sum_{k=1}^m \abs{f(q_k)}^2 + \frac{\lambda_{\text{n}}}{m}\sum_{k=1}^m \norm{\nabla f(q_k) - n_k}^2
\end{equation}
where $\norm{\cdot}$ is the standard euclidean norm in $\Real^3$, $\nabla f(p) \in \Real^3$ is the gradient of $f$ sampled at point $p$. Note that $\nabla f$ is defined in interior of voxels, which is generically where the input points $q_k$ resides. $\gL_{\text{data}}$ is the standard data loss encouraging the zero level to pass through the input points $q_k$, and its normals (defined by gradients of $f$) to coincide with input normals $n_k$. 

The prior, $\gL_{\text{prior}}$, is the main contribution of this work, where we combine two novel losses,
\begin{equation}
    \gL_{\text{prior}} = \lambda_{\text{v}} \gL_{\text{viscosity}} + \lambda_{\text{c}} \gL_{\text{coarea}}
\end{equation}
Intuitively, the viscosity loss optimizes for a smooth Signed Distance Function (SDF) solutions, avoiding auxiliary bad minima of the Eikonal equation, while the coarea loss strives to minimize the area of the zero level surface. Our loss has $4$ hyper-parameters $\lambda_{\text{p}},\lambda_{\text{n}},\lambda_{\text{v}},\lambda_{\text{c}}$. We provide more details on these priors next.

\subsection{Viscosity Loss}\label{ss:viscosity_loss}
The goal of the viscosity loss is to make $f$ approximate an SDF over $\gC$. Given boundary conditions asking $f$ to vanish on some closed compact surface $\gS$, the SDF solves the Eikonal equation PDE, \ie, $\norm{\nabla f(p)}=1$, in a certain well defined sense (viscosity). This motivated some previous work to directly optimize the Eikonal loss \citep{gropp2020implicit,sitzmann2020implicit}
\begin{equation}\label{e:loss_eikonal}
    \gL_{\text{eikonal}} = \int_\gC \Big (\norm{\nabla f(p)}-1\Big )^2 dp
\end{equation}
\begin{wrapfigure}[14]{r}{0.28\textwidth}\vspace{-15pt}
  \begin{center}
    \includegraphics[width=0.25\textwidth]{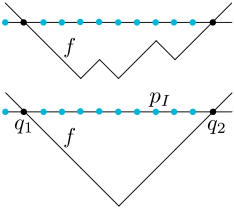}
  \end{center}
  \caption{Two global minimizers of the Eikonal loss over a grid in 1D. Top solution is not an SDF. }\label{fig:eikonal_1d}
\end{wrapfigure}
Unfortunately, the Eikonal loss has many undesirable minima which are not SDFs. Figure \ref{fig:eikonal_1d} shows a 1D example: both depicted solutions (denoted $f$) vanish at the input points $q_1,q_2$ (black points) and globally minimize the Eikonal loss over the grid (grid points are shown in blue). The INR works mentioned above use neural networks for representing $f$ which injects an inductive bias avoiding these bad minima, however on grids, minimizing \eqref{e:loss_eikonal} cannot avoid these solutions. See, \eg, middle column in Figure \ref{fig:teaser}. 

More classical Eikonal solvers do work with grids however use mostly fast marching or sweeping methods \citep{osher1988fronts,sethian1996fast,zhao2005fast,chacon2012fast}. Namely, use a special discretization of the Eikonal equation favoring the viscosity  solution of the Eikonal \cite{rouy1992viscosity}, and update node values according to a moving front \cite{sethian1996fast}. Since this discretization is up-wind (will only propagate values in one direction) and requires choosing the maximal among its solution, its success in adaptation to a loss is not clear. 

We use a different approach to build a loss favoring SDF solutions over grids motivated by the vanishing viscosity method \cite{crandall1983viscosity}. Namely, adding to the Eikonal PDE a small perturbation of the Laplacian of $f$ (denoted by $\Delta f$), \ie, $\norm{\nabla f(p)}-1 - \eps\Delta f(p)=0$, makes the PDE semi-linear elliptic \citep{calder2018lecture}, and hence with suitable boundary conditions it is uniquely solvable inside $\gS$ with a smooth solution, approaching the viscosity positive distance function to the boundary as $\eps\too 0$. Similarly, for $1-\norm{\nabla f(p)} - \eps \Delta f(p)=0$ the solution approaches the negative distance function inside the domain. 
Motivated by the vanishing viscosity principle we suggest the following viscosity loss:
\begin{equation}\label{e:loss_viscosity_eikonal}
\gL_{\text{viscosity}} = \int_\gC \Big((\norm{\nabla f (p)}-1)\mathrm{sign}(f(p)) - \eps \Delta f(p)\Big)^2 dp
\end{equation}
We discretize this loss over the grid $\gG$ by replacing the first order derivatives and second order derivatives with symmetric finite  differences, \ie,
\begin{align*}
    D_x f_I=D_x f_{i,j,k} = \frac{f_{i+1,j,k}-f_{i-1,j,k}}{2h}, \quad D^2_x f_I = D^2_x f_{i,j,k}=\frac{f_{i+1,j,k}-2f_{i,j,k}+f_{i-1,j,k}}{h^2}
\end{align*}
and similarly for $D_y$ and $D_z$. We use these discrete operators to approximate the gradient $\widehat{\nabla} f(p_I) = (D_x f_I, D_y f_I, D_z f_I)$ and Laplacian $\widehat{\Delta}f(p_I) = D_x^2f_I + D_y^2 f_I + D_z^2 f_I$. The discretized viscosity loss now takes the form
\begin{equation}
    \widehat{\gL}_{\text{viscosity}} = \frac{1}{N}\sum_{I} \Big((\|\widehat{\nabla} f (p_I)\|-1)\mathrm{sign}(f(p_I)) - \eps \widehat{\Delta} f(p_I)\Big)^2
\end{equation}

\subsection{Coarea loss}\label{ss:coarea_loss}
The coarea loss is approximating the area of the zero level set, and therefore incorporating it in the optimization pushes the reconstructed surface to be economic in area. 

First, similarly to  \citep{yariv2021volume} we use the centered Laplace CDF
\begin{equation}
   \Psi\beta(s)= \begin{cases}
   \frac{1}{2}\exp\parr{\frac{s}{\beta}} & s\leq 0 \\ 1-\frac{1}{2}\exp\parr{-\frac{s}{\beta}} & s\geq  0
   \end{cases}
\end{equation} to transform the SDF $f$ to a smooth approximation of the indicator function:
\begin{equation}
    \chi_\beta(p)=\Psi\beta (-f(p))
\end{equation}
As $\beta\too 0$, $\chi_\beta$ converges to an indicator function leading to $1$ inside $\gS$ and $0$ outside. The coarea loss is defined as 
\begin{equation}
    \gL_{\text{coarea}} = \int_\gC \norm{\nabla \chi_\beta (p)} dp
\end{equation}
To understand why this loss approximates the area of $\gS$ we can use the coarea formula \citep{rindler2018calculus}:
\begin{equation}\label{e:coarea}
    \int_\gC \norm{\nabla \chi_\beta(p)}dp = \int_{-\infty}^{\infty} \mathrm{area}(\chi_\beta^{-1}(s))ds,
\end{equation}
where $\chi_\beta^{-1}(s)=\set{p\ \vert \ \chi_\beta(p)=s}$ is the preimage of the value $s$. Since $\chi_x(p)\in [0,1]$ the r.h.s.~integral can be restricted to the interval $[0,1]$, and therefore the coarea loss averages the area of the level sets of $\chi_\beta$. Next,  $$\chi_\beta^{-1}(s)= \set{p\ \vert \ \Psi\beta (-f(p)) = s } = \{p\ \vert \ f(p) = -\Psi\beta^{-1} (s) \} = f^{-1}(-\Psi\beta^{-1} (s)),$$
\begin{wrapfigure}[11]{r}{0.28\textwidth}\vspace{-20pt}
  \begin{center}
  \includegraphics[width=0.25\textwidth]{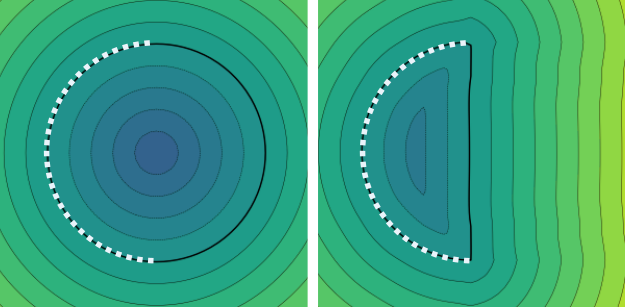}
  \end{center}
  \caption{Reconstruction of a semisphere point cloud (white dots) without (left) and with (right) coarea loss. }\label{fig:coarea_semisphere}
\end{wrapfigure}

which shows that the level set $s\in (0,1)$ of $\chi_\beta$ is the level set $-\Psi\beta^{-1}(s)$ of the SDF $f$. As $\beta\too 0$, $-\Psi\beta^{-1}(s)\too 0$ for all $s\in (0,1)$ (and uniformly in $(\eps,1-\eps)$ for fixed $\eps>0$). Therefore the average of the level set area (\ie, the r.h.s.~of \eqref{e:coarea}) converges to the area of $f^{-1}(0)=\gS$. Figure \ref{fig:teaser} (right) shows how removing the coarea loss introduces an extraneous zero level set, and hence results in an undesired surface part. Figure \ref{fig:coarea_semisphere} shows a comparison of a reconstruction of semisphere with and without coarea. In the experiments section we provide more ablation tests with the coarea and viscosity losses.

To discretize the coarea loss we let $w_I$ denote the centers of grid's voxels, and note that $\nabla \chi_\beta(w_I) = \Phi_\beta(-f(w_I))\nabla f(w_I)$, where 
\begin{equation*}
    \Phi_\beta(s) = \frac{1}{2\beta}\exp\parr{\frac{\abs{s}}{\beta}}
\end{equation*}
is the PDF of the Laplace distribution, and $\nabla f(w_I)$ is computed as a linear combination of the voxel's corner values $f_{I_1},\ldots,f_{I_8}$, see more details in the Appendix. We end up with the discretized loss:
\begin{equation}
    \widehat{\gL}_{\text{coarea}} = \frac{1}{N}\sum_{I}\Phi_\beta(-f(w_I))\norm{\nabla f(w_I)}
\end{equation}
This loss is usually incorporated with a small hyper-parameter $\lambda_{\text{c}}$ with the purpose of eliminating redundant surface parts.

\section{Experiments}

\rebb{In this section we extensively evaluate VisCo grids. First, we evaluate on two standard surface reconstruction benchmarks~\cite{williams2019deep,huang2022surface} (Sec.~\ref{sec:exp_rec}) against a large variety of state-of-the-art methods: Poisson Surface Reconstruction ~\cite{kazhdan2006poisson}, DGP~\cite{williams2019deep}, IGR~\cite{gropp2020implicit}, SIREN~\cite{sitzmann2020implicit}, FFN~\cite{tancik2020fourier}, NSP~\cite{williams2021neural}, PHASE~\cite{lipman2021phase}, GD~\cite{greedydelaunay}, BPA~\cite{BPA}, SPSR~\cite{kazhdan2013screened}, RIMLS~\cite{RIMLS}, SALD~\cite{sald}, IGR~\cite{gropp2020implicitgeometricregularizationforlearningshape}, OccNet~\cite{occupancy}, DeepSDF~\cite{deepsdf}, LIG~\cite{LIG}, Points2Surf~\cite{points2surf}, DSE~\cite{learningdelaunaysurface}, IMLSNet~\cite{liu2021DeepIMLS} and ParseNet~\cite{parsenet}.}
We then perform an ablation study (Sec.~\ref{sec:exp_ablation}), and conduct a detail examination of the main components of the model, namely the viscosity and coarea losses. Finally, we discuss the model’s ability to reconstruct 
sparse point clouds (Sec.~\ref{sec:sparse}) using scans from Stanford 3D Scanning Repository.

\subsection{Surface reconstruction benchmarks}
\label{sec:exp_rec}

\rebb{We next evaluated our model on two benchmarks: Surface Reconstruction Benchmark~\cite{williams2019deep} and Surface Reconstruction from Real-Scans~\cite{huang2022surface}. Each containing challenging object with complex shape and topology. Importantly, we use same hyper-parameters for all meshes of all benchmarks with no extensive hyper-parameter search. }

\begin{figure}[ht!]
    \includegraphics[width=\textwidth]{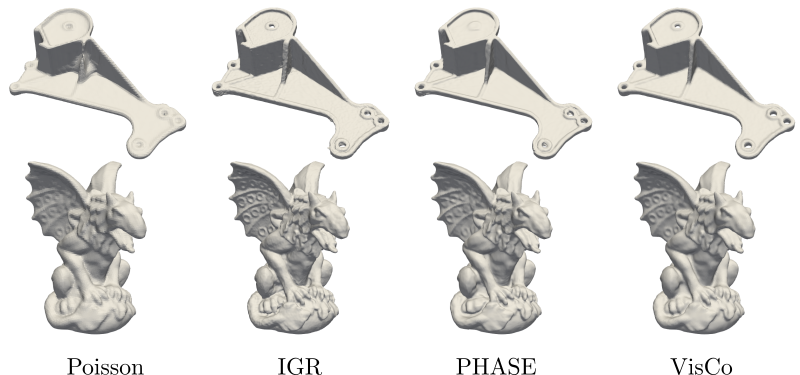}
    \caption{Qualitative results for surface reconstruction \cite{williams2019deep} compared to existing methods. Note how VisCo Grids achieve comparable level of details when compared to other baselines. 
    } \label{fig:recon1}
\end{figure}

\paragraph{Surface Reconstruction Benchmark}
This benchmark~\cite{williams2019deep}
consists of 5 noisy range scans, each containing point cloud and normal data. 
We evaluate our method against current state of the art methods on this benchmark: Deep Geometric Prior (DGP)~\cite{williams2019deep}, Implicit Geometric Regularization (IGR)~\cite{gropp2020implicit}, SIREN~\cite{sitzmann2020implicit}, Fourier Feature Networks (FFN)~\cite{tancik2020fourier}, NSP~\cite{williams2021neural} and PHASE~\cite{lipman2021phase}. We additionally compare to the classical method of Poisson Surface Reconstruction ~\cite{kazhdan2006poisson}.
Quantitative results are summarized in Table~\ref{tab:dgp}. We report the Chamfer ($d_C$) and Hausdorff ($d_H$) distances between the reconstructed meshes and the ground-truth point clouds. Furthermore, we report their corresponding one sided distances ($d_H^\too$ and $d_C^\too$) between the reconstructed meshes and the input noisy point cloud. 
Representative qualitative results are shown in Figure~\ref{fig:recon1}. Note that we achieve comparable results to the current state-of-the-art INR methods.

\begin{figure}[ht!]
    \includegraphics[width=\textwidth]{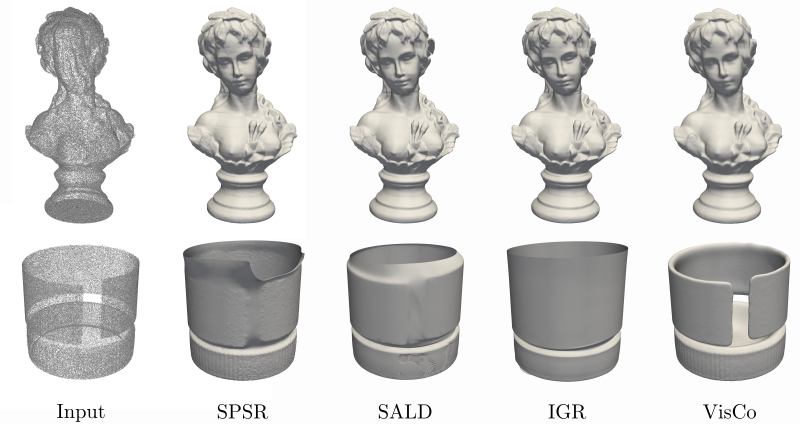}
    \caption{\rebb{Qualitative results for surface reconstruction of real objects \cite{huang2022surface} compared to existing methods. Note how VisCo Grids does not over-extend the surface in the bottom row example. The competing methods meshes were provided by the benchmark organizers.}
    } \label{fig:recon2}
\end{figure}

\rebb{
\paragraph{Surface Reconstruction from Real-Scans} This benchmark~\cite{huang2022surface} consists of 21 noisy range scans of real objects. We evaluate our method against: GD~\cite{greedydelaunay}, BPA~\cite{BPA}, SPSR~\cite{kazhdan2013screened}, RIMLS~\cite{RIMLS}, SALD~\cite{sald}, IGR~\cite{gropp2020implicitgeometricregularizationforlearningshape}, OccNet~\cite{occupancy}, DeepSDF~\cite{deepsdf}, LIG~\cite{LIG}, Points2Surf~\cite{points2surf}, DSE~\cite{learningdelaunaysurface}, IMLSNet~\cite{liu2021DeepIMLS} and ParseNet~\cite{parsenet}. Quantitative results are summarized in Table~\ref{tab:bench2}. We report Chamfer Distance ($d_C$), F-score, Normal Consistency Score (NCS)~\cite{occupancy}, and Neural Feature Similarity (NFS)~\cite{huang2022surface} distances between the reconstructed meshes and the ground-truth point clouds. Furthermore, we report the one sided distances ($d_H^\too$ and $d_C^\too$) between the reconstructed meshes and the input noisy point cloud. Representative qualitative results are shown in Figure~\ref{fig:recon2}. Note that we achieve 1-st to 3-rd place across all categories, with top F-score. 
}

\begin{table}[h!]\scriptsize	
    \centering

    \begin{tabular}{c|c|c|c|c|c|c|c|c|c|c} 
        \multicolumn{2}{c|}{} & Poisson & DGP & IGR & SIREN & FFN & NSP & PHASE & \textbf{Ours \tiny{(30 mins)}} & \textbf{Ours \tiny{(8 mins)}} \\ \midrule
            \bottomrule

        \multirow{4}{*}{Anchor} 
            & $d_C$ & 0.60 & 0.33 & 0.22 & 0.32 & 0.31 & 0.22 & \textbf{0.21} & \textbf{0.21}  & 0.28 \\
            & $d_H$ & 14.89 & 8.82 & 4.71 & 8.19 & 4.49 & 4.65 & 4.29 & \textbf{3.00} & 5.69 \\
            & $d_C^\too$ & 0.60 & 0.08 & 0.12 & 0.10 & 0.10 & 0.11 & 0.09 & 0.15 & 0.15\\
            & $d_H^\too$ & 14.89 & 2.79 & 1.32 & 2.43 & 0.10 & 1.11 & 1.23 & 1.07 & 1.15\\ \hline
        \multirow{4}{*}{Daratech}
            & $d_C$ & 0.44 & 0.20 & 0.25 & 0.21 & 0.34 & 0.21 & \textbf{0.18} & 0.26  & 0.25 \\
            & $d_H$ & 7.24 & 3.14 & 4.01 & 4.30 & 5.97 & 4.35 & \textbf{2.92} & 4.06  & 4.15 \\
            & $d_C^\too$ & 0.44 & 0.04 & 0.08 & 0.09 & 0.10 & 0.08 & 0.08 & 0.14 & 0.13  \\
            & $d_H^\too$ & 7.24 & 1.89 & 1.59 & 1.77 & 0.10 & 1.14 & 1.80 & 1.76 & 1.78 \\ \hline
        \multirow{4}{*}{DC}
            & $d_C$ & 0.27 &0.18 & 0.17 & 0.15 & 0.20 & \textbf{0.14} & 0.15 & 0.15  & 0.15  \\
            & $d_H$ & 3.10 & 4.31 & 2.22 & 2.18 & 2.87 & \textbf{1.35} & 2.52 & 2.22  & 2.23 \\
            & $d_C^\too$ & 0.27 & 0.04 & 0.09 & 0.06 & 0.10 & 0.06 & 0.05 & 0.09  & 0.09 \\
            & $d_H^\too$ & 3.10 & 2.53 & 2.61 & 2.76 & 0.12 & 2.75 & 2.78 & 2.76 & 2.78 \\ \hline
        \multirow{4}{*}{Gargoyle}
            & $d_C$ & 0.26 &0.21 & \textbf{0.16} & 0.17 & 0.22 & \textbf{0.16} & \textbf{0.16} & 0.17  & 0.17 \\
            & $d_H$ & 6.8 & 5.98 & 3.52 & 4.64 & 5.04 & 3.20 & \textbf{3.14} & 4.40  & 4.45 \\
            & $d_C^\too$ & 0.26 & 0.06 & 0.06 & 0.08 & 0.09 & 0.08 & 0.07 & 0.11 & 0.11 \\
            & $d_H^\too$ &  6.80 & 3.41 & 0.81 & 0.91 & 0.09 & 2.75 & 1.09 & 0.96 & 0.98 \\ \hline
        \multirow{4}{*}{Lord Quas}
            & $d_C$ & 0.20 & 0.14 & 0.12 & 0.17 & 0.35 & 0.12 & \textbf{0.11} & 0.12  & 0.13 \\
            & $d_H$ & 4.61 & 3.67 & 1.17 & 0.82 & 3.90 & \textbf{0.69} & 0.96 & 1.06  & 1.14\\
            & $d_C^\too$ & 0.20 & 0.04 & 0.07 & 0.12 & 0.06 & 0.05 & 0.04 & 0.07 & 0.07\\
            & $d_H^\too$ & 4.61 & 2.03 & 0.98 & 0.76 & 0.06 & 0.62 & 0.96 & 0.64 & 0.68\\ 
            \bottomrule

    \end{tabular} \vspace{10pt}
    \caption{Surface reconstruction results on the benchmark of \cite{williams2019deep}. We show reconstruction results for each model for our method at 256 grid resolution with 30 minute and 8 minute time budget. We also show results from comparative methods. Bold numbers signify top performance.  We report Chamfer and Hausdorff distances using ground truth scans ($d_C$, $d_H$) and input scans ($d_C^\too$, $d_H^\too$). Note that VisCo Grids achieve comparable results to SOTA INRs, and even matches it in terms of Chamfer distance in 3 out of 5 meshes.
    }
    \label{tab:dgp}
\end{table}

\begin{table}[h!]\scriptsize	
    \centering
    \begin{tabular}{c|l|c|c|c|c} 
        Prior & Method & $d_C$ $(\times10^{-2}) \downarrow$ & F-score $(\%) \uparrow$ & NCS $(\times10^{-2}) \uparrow$ & NFS $(\times10^{-2}) \uparrow$ \\ \midrule
         \multirow{2}{*}{Triangulation-based}   & GD \cite{greedydelaunay}                                              & 31.72                  & 87.51                  & 88.86                  & 82.20                        \\
                                                        & BPA \cite{BPA}                                                        & 40.37                  & 80.95                  & 87.56                  & 68.69                        \\\cline{1-2}
    \multirow{3}{*}{Smoothness}            & SPSR \cite{kazhdan2013screened}                                       & \textbf{31.05}         & \underline{87.74}         & \underline{{94.94}} & \textbf{89.38}               \\
                                                        & RIMLS \cite{RIMLS}                                                    & 32.80                  & 87.05                  & 91.97                  & 85.19   \\
                                               & \textbf{Ours}                                                     & \phantom{xxxxx}32.11 ($3^{rd}$)                 & \phantom{xxxxx}\textbf{88.52}   ($1^{st}$)                & \phantom{xxxxx}94.20  ($3^{rd}$)                 & \phantom{xxxxx}\underline{89.16} ($2^{rd}$)          
                                            \\\cline{1-2}
    \multirow{2}{*}{Modeling}                & SALD \cite{sald}                                                      & \underline{{31.13}} & {{87.72}} & 94.68                  & 86.86                        \\
                                                        & IGR \cite{gropp2020implicitgeometricregularizationforlearningshape}   & 32.70                  & 87.18                  & \textbf{95.99}         & {{89.10}}       \\\cline{1-2}
    \multirow{2}{*}{Learning Semantics}     & OccNet \cite{occupancy}                                               & 232.71                 & 17.11                  & 80.96                  & 39.70                        \\
                                                        & DeepSDF \cite{deepsdf}                                                & 263.92                 & 19.83                  & 77.95                  & 40.95                        \\\cline{1-2}
    \multirow{2}{*}{Local Learning}         & LIG \cite{LIG}                                                        & 48.75                  & 83.76                  & 92.57                  & 81.48                        \\
                                                        & Points2Surf \cite{points2surf}                                        & 48.93                  & 80.89                  & 89.52                  & 81.83                        \\\cline{1-2}
     \multirow{3}{*}{Hybird}                                                   & DSE \cite{learningdelaunaysurface}                                    & 32.16                  & 86.88                  & 87.20                  & 76.81                        \\
                      & IMLSNet \cite{liu2021DeepIMLS}                                        & 38.46                  & 82.44                  & 93.31                  & 85.30                        \\
                                                        & ParseNet \cite{parsenet}                                              & 149.96                 & 38.92                  & 81.51                  & 45.67                        \\\hline

    \end{tabular} \vspace{10pt}
    \caption{\rebb{Surface reconstruction results on the 20 real-scanned benchmark~\cite{huang2022surface} meshes. We report Chamfer Distance ($d_C$), F-score, Normal Consistency Score (NCS)~\cite{occupancy}, and Neural Feature Similarity (NFS)~\cite{huang2022surface}. Methods are grouped according to surface geometry priors, as originally defined in the benchmark. Our method achieves top F-score and 1-st to 3-rd place in all scores.}}
    \label{tab:bench2}
\end{table}

\subsection{Ablation study}
\label{sec:exp_ablation}

\begin{figure}[ht!]
    \includegraphics[width=\textwidth]{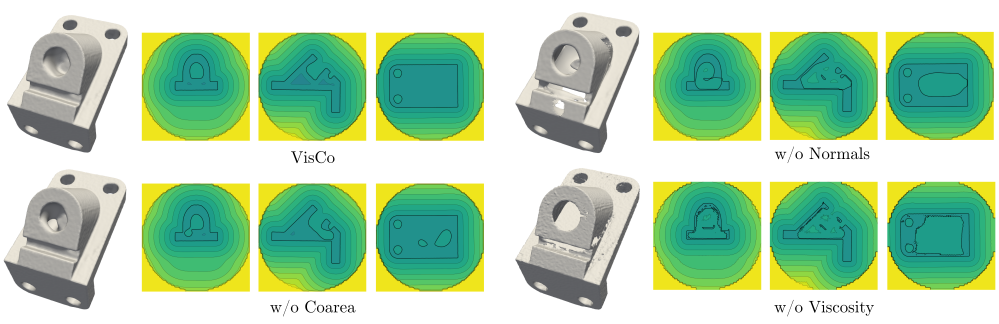} \vspace{-20pt}
    \caption{Ablation for the main components of our method. Removing elements of our loss leads to subpar reconstructions. We can observe these artifacts in the level sets shown in this figure. Removing viscosity results in discontinuities in the final surface, while no coarea produces excess surface area.} \label{fig:abl}
    \vspace{-15pt}
\end{figure}

\begin{table}[h!]\scriptsize	
    \centering
    \begin{tabular}{c|c|c|c|c|c}
        \multicolumn{2}{c|}{} & Baseline & w/o normals & w/o viscosity & w/o coarea \\ \hline
        \multirow{4}{*}{Anchor}
            & $d_C$ & \textbf{0.21} & 0.61 & 0.55 & 0.72 \\
            & $d_H$ & \textbf{3.00} & 7.82 & 10.83 & 10.24 \\
            & $d_C^\too$ & 0.15 & 0.37 & 0.27 & 0.36 \\
            & $d_H^\too$ & 1.07 & 7.84 & 1.44 & 9.68 \\ \hline
        \multirow{4}{*}{Daratech}
            & $d_C$ & 0.26 & 0.24 & 0.24 & \textbf{0.23} \\
            & $d_H$ & 4.06 & 4.2 & 4.3 & \textbf{2.19} \\
            & $d_C^\too$ & 0.14 & 0.13 & 0.12 & 0.13 \\
            & $d_H^\too$ & 1.76 & 2.69 & 1.77 & 1.77 \\ \hline
        \multirow{4}{*}{DC}
            & $d_C$ & \textbf{0.15} & \textbf{0.15} & \textbf{0.15} & 0.34 \\
            & $d_H$ & \textbf{2.22} & 2.24 & 2.24 & 6.58 \\
            & $d_C^\too$ & 0.09 & 0.08 & 0.08 & 0.16 \\
            & $d_H^\too$ & 2.76 & 2.76 & 2.79 & 2.82 \\ \hline
        \multirow{4}{*}{Gargoyle}
            & $d_C$ & \textbf{0.17} & 0.58 & 0.47 & 0.59 \\
            & $d_H$ & \textbf{4.40} & 6.32 & 10.38 & 6.35 \\
            & $d_C^\too$ & 0.11 & 0.07 & 0.26 & 0.38 \\
            & $d_H^\too$ & 0.96 & 2.39 & 1.34 & 1.25 \\ \hline
        \multirow{4}{*}{Lord Quas}
            & $d_C$ & \textbf{0.12} & 0.12 & 0.12 & 0.58 \\
            & $d_H$ & 1.06 & 1.38 & \textbf{1.04} & 6.05 \\
            & $d_C^\too$ & 0.07 & 0.37 & 0.06 & 0.32 \\
            & $d_H^\too$ & 0.64 & 0.69 & 0.64 & 3.73 \\ \hline %
            
    \end{tabular} \vspace{5pt}
    \caption{Ablations study. We show the contribution of each component of VisCo Grids. Baseline is the full method. The remaining columns correspond to optimizing without normal loss, viscosity loss and coarea loss, respectively. We show results for each mesh of the benchmark \cite{williams2019deep}. The results justify the use of the different components in VisCo Grids.}
    \label{tab:ablations}
\end{table}

We provide an ablation study of the main components of our model in Table~\ref{tab:ablations} and Figure~\ref{fig:abl}. Specifically we compared with the following alternatives: i) Eikonal loss without the viscosity term that prevents undesirable non-SDF solutions, \ie, $\eps=0$ in \eqref{e:loss_viscosity_eikonal}; ii) removing the coarea loss enforcing minimal surface area, \ie, $\lambda_{\text{c}}=0$;  and iii) removing the normal loss, \ie, $\lambda_{\text{n}}=0$. Note that without coarea and viscosity the reconstruction tends to have holes and discontinuities near the surface boundaries. Only combination of all the components results in a good surface reconstruction.

\begin{figure}[ht!]
\vspace{-20pt}
    \includegraphics[width=\textwidth]{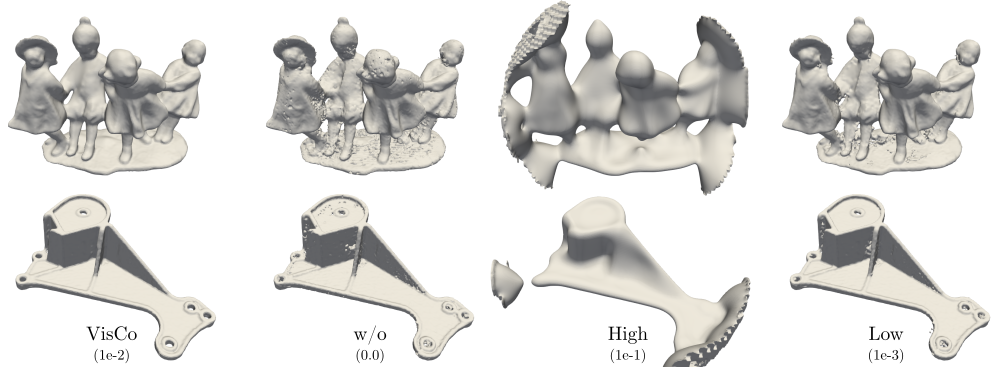} \vspace{-10pt}
    \caption{Viscosity loss ablation. Setting a high viscosity loss parameter, $\eps$, leads to oversmoothing. In contrast, setting it too low leads to noise and discontinuities in the surfaces, similarly to removing it by setting $\eps=0$.} \vspace{-20pt} \label{fig:viscosity}
\end{figure}

\paragraph{Learning with viscosity.} We further provide a more in depth discussion of the proposed viscosity loss. Figure~\ref{fig:viscosity} compares reconstructions with different levels of the viscosity parameter, \ie, $\eps$ in \eqref{e:loss_viscosity_eikonal}. As can be inspected from this figure, viscosity affects the smoothness of the reconstructed surface. For a low viscosity parameter the zero level sets become noisy. This can be explained by the viscosity eikonal loss (\ie, \eqref{e:loss_viscosity_eikonal}) becoming numerically too close to the eikonal loss in \eqref{e:loss_eikonal} and the solution deviates from the viscosity SDF solution. This leads to artifacts across the surface, similar to the limit case ($\eps=0$) where only the eikonal loss is used, see the second column from the left. For a high viscosity parameter, and as expected with the addition of a non-vanishing Laplacian term, the surface becomes over-smoothed.

\begin{figure}[ht!]
    \includegraphics[width=\textwidth]{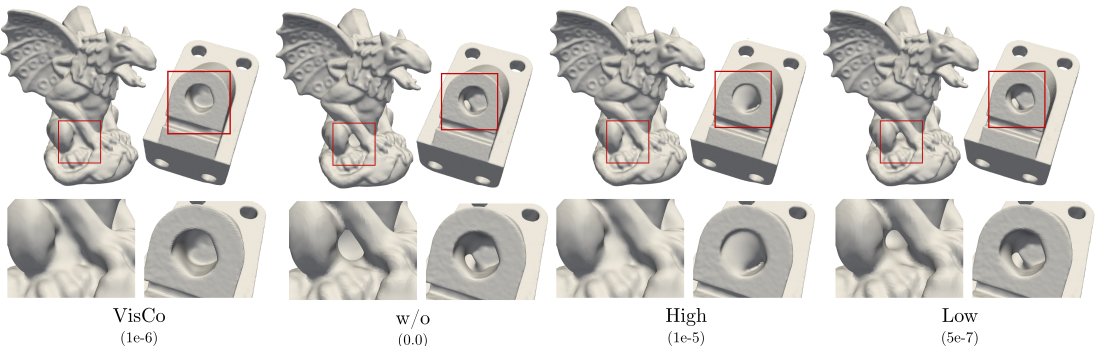} \vspace{-10pt}
    \caption{Coarea loss ablation. Coarea loss favors solution with low surface area. Here, note how larger coarea weight tends to fill in the cavities and close the gaps in the shape. In contrast, very low weight fails to drive the optimization procedure towards a solution with smaller surface area. } \label{fig:coarea}
    \vspace{-10pt}
\end{figure}

\paragraph{Learning with coarea. } Similarly, we also provide an analysis of the proposed coarea loss. 
In Figure~\ref{fig:coarea} we show the effect of changing the parameter weight of the coarea loss, $\lambda_{\text{c}}$. As can be observed in the figure, a low coarea weight leads to the presence of excessive surface area in the reconstruction. In contrast, a high weight will strive to minimize the surface area. In a sense, the coarea serves a surface tension parameter; stronger tension will ignore points, weaker tension will overfit. 

\begin{figure}[ht!]
\centering
    \includegraphics[width=0.8\textwidth]{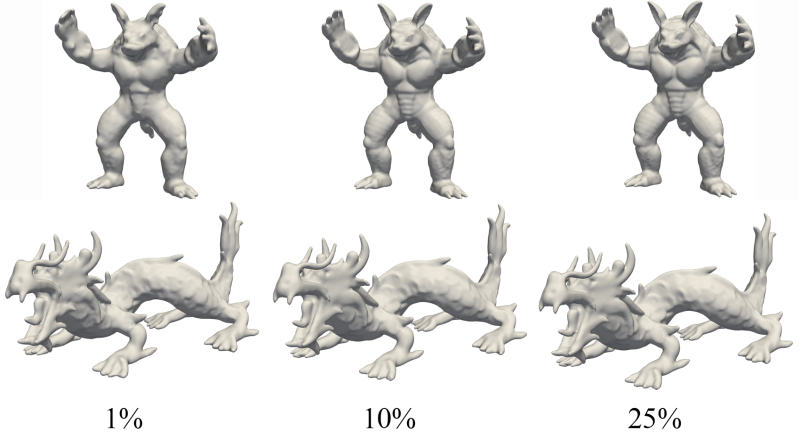}\vspace{-10pt}
    \caption{VisCo reconstructions from sparse point cloud inputs. }\vspace{-10pt} \label{fig:sparse}
\end{figure}

\subsection{Reconstructing from sparse point clouds}
\label{sec:sparse}
We now evaluate our model ability to reconstruct surfaces in the challenging case of sparse input point clouds. For this experiment we use point clouds from the Stanford 3D Scanning Repository and downsample them at different levels: 1\%, 10\%, and 25\%. In Figure~\ref{fig:sparse} we visualize the VisCo reconstructions. Note that VisCo can reconstruct the shapes even with sparse input. This provides a further validation for the proposed geometrical priors.\vspace{-5pt}

\section{Conclusions}\vspace{-5pt}

We introduced VisCo Grids, a novel grid-based surface reconstruction approach that leverages two novel geometrically motivated priors: viscosity and coarea. We advocate VisCo's prior for grid functions as an alternative to the implicit inductive bias of implicit neural representations for the task of surface reconstruction. One important limitation of our method, shared by all grid methods, is that its degrees of freedom, namely nodes' location, are set a-priori. In contrast, using non-linear function spaces, such as neural networks, allows a more flexible usage of the degrees of freedoms in the model and can adjust to areas with more detail. Nevertheless, we still believe that grid functions and direct priors, incorporated with modern computing power, are valuable add-ons to the surface reconstruction toolbox, providing few clear benefits over neural networks, such as a well-understood control over surface properties, instant inference time, and faster training. 

\paragraph{Acknowledgements.}
We thank the anonymous reviewers for their constructive comments. LY was supported by a grant from Israel CHE Program for Data Science Research Centers.

\paragraph{Social impact.} We don't see any immediate negative societal impact from our work. However, we acknowledge that high quality 3D reconstruction in general can be used in malicious settings.

{
\bibliographystyle{plainnat}
\bibliography{ms}

\begin{thebibliography}{51}
\providecommand{\natexlab}[1]{#1}
\providecommand{\url}[1]{\texttt{#1}}
\expandafter\ifx\csname urlstyle\endcsname\relax
  \providecommand{\doi}[1]{doi: #1}\else
  \providecommand{\doi}{doi: \begingroup \urlstyle{rm}\Url}\fi

\bibitem[{Alex Yu and Sara Fridovich-Keil} et~al.(2021){Alex Yu and Sara
  Fridovich-Keil}, Tancik, Chen, Recht, and Kanazawa]{yu2021plenoxels}
{Alex Yu and Sara Fridovich-Keil}, Matthew Tancik, Qinhong Chen, Benjamin
  Recht, and Angjoo Kanazawa.
\newblock Plenoxels: Radiance fields without neural networks, 2021.

\bibitem[Amenta et~al.(1998)Amenta, Bern, and Kamvysselis]{amenta1998new}
Nina Amenta, Marshall Bern, and Manolis Kamvysselis.
\newblock A new voronoi-based surface reconstruction algorithm.
\newblock In \emph{Proceedings of the 25th annual conference on Computer
  graphics and interactive techniques}, pages 415--421, 1998.

\bibitem[Atzmon and Lipman(2019)]{atzmon2019sal}
Matan Atzmon and Yaron Lipman.
\newblock Sal: Sign agnostic learning of shapes from raw data.
\newblock \emph{arXiv preprint arXiv:1911.10414}, 2019.

\bibitem[Atzmon and Lipman(2021{\natexlab{a}})]{atzmon2021sald}
Matan Atzmon and Yaron Lipman.
\newblock {SALD:} sign agnostic learning with derivatives.
\newblock In \emph{9th International Conference on Learning Representations,
  {ICLR} 2021}, 2021{\natexlab{a}}.

\bibitem[Atzmon and Lipman(2021{\natexlab{b}})]{sald}
Matan Atzmon and Yaron Lipman.
\newblock Sald: Sign agnostic learning with derivatives.
\newblock In \emph{International Conference on Learning Representations},
  2021{\natexlab{b}}.

\bibitem[Ben-Shabat et~al.(2022)Ben-Shabat, Koneputugodage, and
  Gould]{ben2022digs}
Yizhak Ben-Shabat, Chamin~Hewa Koneputugodage, and Stephen Gould.
\newblock Digs: Divergence guided shape implicit neural representation for
  unoriented point clouds.
\newblock In \emph{Proceedings of the IEEE/CVF Conference on Computer Vision
  and Pattern Recognition}, pages 19323--19332, 2022.

\bibitem[Berger et~al.(2017)Berger, Tagliasacchi, Seversky, Alliez, Guennebaud,
  Levine, Sharf, and Silva]{berger2017survey}
Matthew Berger, Andrea Tagliasacchi, Lee~M Seversky, Pierre Alliez, Gael
  Guennebaud, Joshua~A Levine, Andrei Sharf, and Claudio~T Silva.
\newblock A survey of surface reconstruction from point clouds.
\newblock In \emph{Computer Graphics Forum}, volume~36, pages 301--329. Wiley
  Online Library, 2017.

\bibitem[Bernardini et~al.(1999)Bernardini, Mittleman, Rushmeier, Silva, and
  Taubin]{BPA}
Fausto Bernardini, Joshua Mittleman, Holly Rushmeier, Cl{\'a}udio Silva, and
  Gabriel Taubin.
\newblock The ball-pivoting algorithm for surface reconstruction.
\newblock \emph{IEEE transactions on visualization and computer graphics},
  5\penalty0 (4):\penalty0 349--359, 1999.

\bibitem[Calder(2018)]{calder2018lecture}
Jeff Calder.
\newblock Lecture notes on viscosity solutions.
\newblock \emph{Lecture notes}, 2018.

\bibitem[Carr et~al.(2001)Carr, Beatson, Cherrie, Mitchell, Fright, McCallum,
  and Evans]{carr2001reconstruction}
Jonathan~C Carr, Richard~K Beatson, Jon~B Cherrie, Tim~J Mitchell, W~Richard
  Fright, Bruce~C McCallum, and Tim~R Evans.
\newblock Reconstruction and representation of 3d objects with radial basis
  functions.
\newblock In \emph{Proceedings of the 28th annual conference on Computer
  graphics and interactive techniques}, pages 67--76, 2001.

\bibitem[Chacon and Vladimirsky(2012)]{chacon2012fast}
Adam Chacon and Alexander Vladimirsky.
\newblock Fast two-scale methods for eikonal equations.
\newblock \emph{SIAM Journal on Scientific Computing}, 34\penalty0
  (2):\penalty0 A547--A578, 2012.

\bibitem[Chambolle et~al.(2010)Chambolle, Novaga, Cremers, and
  Pock]{Chambolle10anintroduction}
A.~Chambolle, M.~Novaga, D.~Cremers, and T.~Pock.
\newblock An introduction to total variation for image analysis.
\newblock In \emph{in Theoretical Foundations and Numerical Methods for Sparse
  Recovery, De Gruyter}, 2010.

\bibitem[Chen and Zhang(2019)]{chen2019learning}
Zhiqin Chen and Hao Zhang.
\newblock Learning implicit fields for generative shape modeling.
\newblock In \emph{Proceedings of the IEEE/CVF Conference on Computer Vision
  and Pattern Recognition}, pages 5939--5948, 2019.

\bibitem[Cohen-Steiner and Da(2004)]{greedydelaunay}
David Cohen-Steiner and Frank Da.
\newblock A greedy delaunay-based surface reconstruction algorithm.
\newblock \emph{The visual computer}, 20\penalty0 (1):\penalty0 4--16, 2004.

\bibitem[Crandall and Lions(1983)]{crandall1983viscosity}
Michael~G Crandall and Pierre-Louis Lions.
\newblock Viscosity solutions of hamilton-jacobi equations.
\newblock \emph{Transactions of the American mathematical society},
  277\penalty0 (1):\penalty0 1--42, 1983.

\bibitem[Erler et~al.(2020{\natexlab{a}})Erler, Guerrero, Ohrhallinger, Mitra,
  and Wimmer]{erler2020points2surf}
Philipp Erler, Paul Guerrero, Stefan Ohrhallinger, Niloy~J Mitra, and Michael
  Wimmer.
\newblock Points2surf learning implicit surfaces from point clouds.
\newblock In \emph{European Conference on Computer Vision}, pages 108--124.
  Springer, 2020{\natexlab{a}}.

\bibitem[Erler et~al.(2020{\natexlab{b}})Erler, Guerrero, Ohrhallinger, Mitra,
  and Wimmer]{points2surf}
Philipp Erler, Paul Guerrero, Stefan Ohrhallinger, Niloy~J Mitra, and Michael
  Wimmer.
\newblock Points2surf learning implicit surfaces from point clouds.
\newblock In \emph{European Conference on Computer Vision}, pages 108--124.
  Springer, 2020{\natexlab{b}}.

\bibitem[Gropp et~al.(2020{\natexlab{a}})Gropp, Yariv, Haim, Atzmon, and
  Lipman]{gropp2020implicit}
Amos Gropp, Lior Yariv, Niv Haim, Matan Atzmon, and Yaron Lipman.
\newblock Implicit geometric regularization for learning shapes.
\newblock \emph{arXiv preprint arXiv:2002.10099}, 2020{\natexlab{a}}.

\bibitem[Gropp et~al.(2020{\natexlab{b}})Gropp, Yariv, Haim, Atzmon, and
  Lipman]{gropp2020implicitgeometricregularizationforlearningshape}
Amos Gropp, Lior Yariv, Niv Haim, Matan Atzmon, and Yaron Lipman.
\newblock Implicit geometric regularization for learning shapes.
\newblock In \emph{Proceedings of Machine Learning and Systems 2020}, pages
  3569--3579. 2020{\natexlab{b}}.

\bibitem[Groueix et~al.(2018)Groueix, Fisher, Kim, Russell, and
  Aubry]{groueix2018papier}
Thibault Groueix, Matthew Fisher, Vladimir~G Kim, Bryan~C Russell, and Mathieu
  Aubry.
\newblock A papier-m{\^a}ch{\'e} approach to learning 3d surface generation.
\newblock In \emph{Proceedings of the IEEE conference on computer vision and
  pattern recognition}, pages 216--224, 2018.

\bibitem[Hanocka et~al.(2020)Hanocka, Metzer, Giryes, and
  Cohen-Or]{hanocka2020point2mesh}
Rana Hanocka, Gal Metzer, Raja Giryes, and Daniel Cohen-Or.
\newblock Point2mesh: A self-prior for deformable meshes.
\newblock \emph{arXiv preprint arXiv:2005.11084}, 2020.

\bibitem[Huang et~al.(2022)Huang, Wen, Wang, Ren, and Jia]{huang2022surface}
Zhangjin Huang, Yuxin Wen, Zihao Wang, Jinjuan Ren, and Kui Jia.
\newblock Surface reconstruction from point clouds: A survey and a benchmark.
\newblock \emph{arXiv preprint arXiv:2205.02413}, 2022.

\bibitem[Jiang et~al.(2020)Jiang, Sud, Makadia, Huang, Nie{\ss}ner, and
  Funkhouser]{LIG}
Chiyu Jiang, Avneesh Sud, Ameesh Makadia, Jingwei Huang, Matthias Nie{\ss}ner,
  and Thomas Funkhouser.
\newblock Local implicit grid representations for 3d scenes.
\newblock In \emph{Proceedings of the IEEE/CVF Conference on Computer Vision
  and Pattern Recognition}, pages 6001--6010, 2020.

\bibitem[Kazhdan and Hoppe(2013)]{kazhdan2013screened}
Michael Kazhdan and Hugues Hoppe.
\newblock Screened poisson surface reconstruction.
\newblock \emph{ACM Transactions on Graphics (ToG)}, 32\penalty0 (3):\penalty0
  1--13, 2013.

\bibitem[Kazhdan et~al.(2006)Kazhdan, Bolitho, and Hoppe]{kazhdan2006poisson}
Michael Kazhdan, Matthew Bolitho, and Hugues Hoppe.
\newblock Poisson surface reconstruction.
\newblock In \emph{Proceedings of the fourth Eurographics symposium on Geometry
  processing}, volume~7, 2006.

\bibitem[Kingma and Ba(2014)]{kingma2014adam}
Diederik~P Kingma and Jimmy Ba.
\newblock Adam: A method for stochastic optimization.
\newblock \emph{arXiv preprint arXiv:1412.6980}, 2014.

\bibitem[Lindell et~al.(2021)Lindell, Van~Veen, Park, and
  Wetzstein]{lindell2021bacon}
David~B Lindell, Dave Van~Veen, Jeong~Joon Park, and Gordon Wetzstein.
\newblock Bacon: Band-limited coordinate networks for multiscale scene
  representation.
\newblock \emph{arXiv preprint arXiv:2112.04645}, 2021.

\bibitem[Lipman(2021)]{lipman2021phase}
Yaron Lipman.
\newblock Phase transitions, distance functions, and implicit neural
  representations.
\newblock \emph{arXiv preprint arXiv:2106.07689}, 2021.

\bibitem[Liu et~al.(2021)Liu, Guo, Pan, Wang, Tong, and Liu]{liu2021DeepIMLS}
Shi-Lin Liu, Hao-Xiang Guo, Hao Pan, Peng-Shuai Wang, Xin Tong, and Yang Liu.
\newblock Deep implicit moving least-squares functions for 3d reconstruction.
\newblock In \emph{Proceedings of the IEEE/CVF Conference on Computer Vision
  and Pattern Recognition}, pages 1788--1797, 2021.

\bibitem[Mescheder et~al.(2019{\natexlab{a}})Mescheder, Oechsle, Niemeyer,
  Nowozin, and Geiger]{mescheder2019occupancy}
Lars Mescheder, Michael Oechsle, Michael Niemeyer, Sebastian Nowozin, and
  Andreas Geiger.
\newblock Occupancy networks: Learning 3d reconstruction in function space.
\newblock In \emph{Proceedings of the IEEE Conference on Computer Vision and
  Pattern Recognition}, pages 4460--4470, 2019{\natexlab{a}}.

\bibitem[Mescheder et~al.(2019{\natexlab{b}})Mescheder, Oechsle, Niemeyer,
  Nowozin, and Geiger]{occupancy}
Lars Mescheder, Michael Oechsle, Michael Niemeyer, Sebastian Nowozin, and
  Andreas Geiger.
\newblock Occupancy networks: Learning 3d reconstruction in function space.
\newblock In \emph{Proceedings of the IEEE Conference on Computer Vision and
  Pattern Recognition}, pages 4460--4470, 2019{\natexlab{b}}.

\bibitem[Mildenhall et~al.(2020)Mildenhall, Srinivasan, Tancik, Barron,
  Ramamoorthi, and Ng]{mildenhall2020nerf}
Ben Mildenhall, Pratul~P Srinivasan, Matthew Tancik, Jonathan~T Barron, Ravi
  Ramamoorthi, and Ren Ng.
\newblock Nerf: Representing scenes as neural radiance fields for view
  synthesis.
\newblock In \emph{European conference on computer vision}, pages 405--421.
  Springer, 2020.

\bibitem[M\"uller et~al.(2022)M\"uller, Evans, Schied, and
  Keller]{mueller2022instant}
Thomas M\"uller, Alex Evans, Christoph Schied, and Alexander Keller.
\newblock Instant neural graphics primitives with a multiresolution hash
  encoding.
\newblock \emph{ACM Trans. Graph.}, 41\penalty0 (4):\penalty0 102:1--102:15,
  July 2022.
\newblock \doi{10.1145/3528223.3530127}.
\newblock URL \url{https://doi.org/10.1145/3528223.3530127}.

\bibitem[Osher and Sethian(1988)]{osher1988fronts}
Stanley Osher and James~A Sethian.
\newblock Fronts propagating with curvature-dependent speed: Algorithms based
  on hamilton-jacobi formulations.
\newblock \emph{Journal of computational physics}, 79\penalty0 (1):\penalty0
  12--49, 1988.

\bibitem[{\"O}ztireli et~al.(2009){\"O}ztireli, Guennebaud, and Gross]{RIMLS}
A~Cengiz {\"O}ztireli, Gael Guennebaud, and Markus Gross.
\newblock Feature preserving point set surfaces based on non-linear kernel
  regression.
\newblock In \emph{Computer Graphics Forum}, volume~28, pages 493--501. Wiley
  Online Library, 2009.

\bibitem[Park et~al.(2019{\natexlab{a}})Park, Florence, Straub, Newcombe, and
  Lovegrove]{deepsdf}
Jeong~Joon Park, Peter Florence, Julian Straub, Richard Newcombe, and Steven
  Lovegrove.
\newblock Deepsdf: Learning continuous signed distance functions for shape
  representation.
\newblock In \emph{Proceedings of the IEEE Conference on Computer Vision and
  Pattern Recognition}, pages 165--174, 2019{\natexlab{a}}.

\bibitem[Park et~al.(2019{\natexlab{b}})Park, Florence, Straub, Newcombe, and
  Lovegrove]{park2019deepsdf}
Jeong~Joon Park, Peter Florence, Julian Straub, Richard Newcombe, and Steven
  Lovegrove.
\newblock Deepsdf: Learning continuous signed distance functions for shape
  representation.
\newblock In \emph{Proceedings of the IEEE Conference on Computer Vision and
  Pattern Recognition}, pages 165--174, 2019{\natexlab{b}}.

\bibitem[Peng et~al.(2021)Peng, Jiang, Liao, Niemeyer, Pollefeys, and
  Geiger]{peng2021shape}
Songyou Peng, Chiyu Jiang, Yiyi Liao, Michael Niemeyer, Marc Pollefeys, and
  Andreas Geiger.
\newblock Shape as points: A differentiable poisson solver.
\newblock \emph{Advances in Neural Information Processing Systems}, 34, 2021.

\bibitem[Rakotosaona et~al.(2021)Rakotosaona, Guerrero, Aigerman, Mitra, and
  Ovsjanikov]{learningdelaunaysurface}
Marie-Julie Rakotosaona, Paul Guerrero, Noam Aigerman, Niloy~J Mitra, and Maks
  Ovsjanikov.
\newblock Learning delaunay surface elements for mesh reconstruction.
\newblock In \emph{Proceedings of the IEEE/CVF Conference on Computer Vision
  and Pattern Recognition}, pages 22--31, 2021.

\bibitem[Rindler(2018)]{rindler2018calculus}
Filip Rindler.
\newblock \emph{Calculus of variations}, volume~5.
\newblock Springer, 2018.

\bibitem[Rouy and Tourin(1992)]{rouy1992viscosity}
Elisabeth Rouy and Agnes Tourin.
\newblock A viscosity solutions approach to shape-from-shading.
\newblock \emph{SIAM Journal on Numerical Analysis}, 29\penalty0 (3):\penalty0
  867--884, 1992.

\bibitem[Sethian(1996)]{sethian1996fast}
James~A Sethian.
\newblock A fast marching level set method for monotonically advancing fronts.
\newblock \emph{Proceedings of the National Academy of Sciences}, 93\penalty0
  (4):\penalty0 1591--1595, 1996.

\bibitem[Sharma et~al.(2020)Sharma, Liu, Maji, Kalogerakis, Chaudhuri, and
  M{\v{e}}ch]{parsenet}
Gopal Sharma, Difan Liu, Subhransu Maji, Evangelos Kalogerakis, Siddhartha
  Chaudhuri, and Radom{\'\i}r M{\v{e}}ch.
\newblock Parsenet: A parametric surface fitting network for 3d point clouds.
\newblock In \emph{European Conference on Computer Vision}, pages 261--276.
  Springer, 2020.

\bibitem[Sitzmann et~al.(2020)Sitzmann, Martel, Bergman, Lindell, and
  Wetzstein]{sitzmann2020implicit}
Vincent Sitzmann, Julien Martel, Alexander Bergman, David Lindell, and Gordon
  Wetzstein.
\newblock Implicit neural representations with periodic activation functions.
\newblock \emph{Advances in Neural Information Processing Systems},
  33:\penalty0 7462--7473, 2020.

\bibitem[Takikawa et~al.(2021)Takikawa, Litalien, Yin, Kreis, Loop,
  Nowrouzezahrai, Jacobson, McGuire, and Fidler]{takikawa2021nglod}
Towaki Takikawa, Joey Litalien, Kangxue Yin, Karsten Kreis, Charles Loop, Derek
  Nowrouzezahrai, Alec Jacobson, Morgan McGuire, and Sanja Fidler.
\newblock Neural geometric level of detail: Real-time rendering with implicit
  {3D} shapes.
\newblock 2021.

\bibitem[Tancik et~al.(2020)Tancik, Srinivasan, Mildenhall, Fridovich-Keil,
  Raghavan, Singhal, Ramamoorthi, Barron, and Ng]{tancik2020fourier}
Matthew Tancik, Pratul Srinivasan, Ben Mildenhall, Sara Fridovich-Keil, Nithin
  Raghavan, Utkarsh Singhal, Ravi Ramamoorthi, Jonathan Barron, and Ren Ng.
\newblock Fourier features let networks learn high frequency functions in low
  dimensional domains.
\newblock \emph{Advances in Neural Information Processing Systems},
  33:\penalty0 7537--7547, 2020.

\bibitem[Williams et~al.(2019)Williams, Schneider, Silva, Zorin, Bruna, and
  Panozzo]{williams2019deep}
Francis Williams, Teseo Schneider, Claudio Silva, Denis Zorin, Joan Bruna, and
  Daniele Panozzo.
\newblock Deep geometric prior for surface reconstruction.
\newblock In \emph{Proceedings of the IEEE/CVF Conference on Computer Vision
  and Pattern Recognition}, pages 10130--10139, 2019.

\bibitem[Williams et~al.(2021)Williams, Trager, Bruna, and
  Zorin]{williams2021neural}
Francis Williams, Matthew Trager, Joan Bruna, and Denis Zorin.
\newblock Neural splines: Fitting 3d surfaces with infinitely-wide neural
  networks.
\newblock In \emph{Proceedings of the IEEE/CVF Conference on Computer Vision
  and Pattern Recognition}, pages 9949--9958, 2021.

\bibitem[Yariv et~al.(2020)Yariv, Kasten, Moran, Galun, Atzmon, Ronen, and
  Lipman]{yariv2020multiview}
Lior Yariv, Yoni Kasten, Dror Moran, Meirav Galun, Matan Atzmon, Basri Ronen,
  and Yaron Lipman.
\newblock Multiview neural surface reconstruction by disentangling geometry and
  appearance.
\newblock \emph{Advances in Neural Information Processing Systems},
  33:\penalty0 2492--2502, 2020.

\bibitem[Yariv et~al.(2021)Yariv, Gu, Kasten, and Lipman]{yariv2021volume}
Lior Yariv, Jiatao Gu, Yoni Kasten, and Yaron Lipman.
\newblock Volume rendering of neural implicit surfaces.
\newblock \emph{Advances in Neural Information Processing Systems}, 34, 2021.

\bibitem[Zhao(2005)]{zhao2005fast}
Hongkai Zhao.
\newblock A fast sweeping method for eikonal equations.
\newblock \emph{Mathematics of computation}, 74\penalty0 (250):\penalty0
  603--627, 2005.

\end{thebibliography}
}

\newpage

\renewcommand*\thetable{\Roman{table}}
\renewcommand*\thefigure{\Roman{figure}}
\setcounter{table}{0}
\setcounter{figure}{0}

\section*{Appendix}
\appendix

\section{Computing  $\nabla f(w_I)$}

We will use the notation set in Section 3. For the losses in Eq.~3 in the main paper (the normal term) and Eq.~11 we require computing $\nabla f(w_I)$, where $w_I$ is the center of the voxel of interest. For simplicity we will consider the voxel $[0,h]^3$. The 8 trilinear basis functions for this voxel are \begin{equation}
    \varphi_{abc}(x,y,z) = 
    \frac{1}{h^3}\begin{Bmatrix}
x & \text{if }a=0\\ 
h-x & \text{if }a=1 
\end{Bmatrix}\cdot
\begin{Bmatrix}
y & \text{if }b=0\\ 
h-y & \text{if }b=1 
\end{Bmatrix}\cdot
\begin{Bmatrix}
z & \text{if }c=0\\ 
h-z & \text{if }c=1 
\end{Bmatrix}
\end{equation}
where the corners of the voxel are indexed by $a,b,c\in\set{0,1}$. Given function values at these corner nodes, $f_{abc}$, the trilinear interpolant of $f$ inside the voxel is
\begin{equation}
    f_{abc}(x,y,z) = \sum_{a,b,c\in\set{0,1}}f_{abc} \varphi_{abc}(x,y,z).
\end{equation}
Taking the gradient of this interpolant we get
\begin{equation}
    \nabla f_{abc}(x,y,z) = \sum_{a,b,c\in\set{0,1}}f_{abc} \nabla \varphi_{abc}(x,y,z)
\end{equation}
and for the center voxel point,  $(x,y,z)=\frac{1}{2}(h,h,h)$, we have
\begin{equation}
    \nabla \varphi_{abc}\parr{\frac{h}{2},\frac{h}{2},\frac{h}{2}} = \frac{1}{4h}\brac{(-1)^a, (-1)^b, (-1)^c}.
\end{equation}
Similarly we can compute the gradient at an arbitrary point $(x, y, z)$ inside a voxel.

\section{Implementation Details}\label{s:implementation_details}
For all experiments we follow a coarse-to-fine approach. We start optimizing at a $64\times64\times64$ grid resolution, then scale up to $128\times128\times128$ and finish at $256\times256\times256$. At each scale up we initialize the higher resolution grid values, $f_I$, by using trilinear interpolation within the voxels of the coarser grid. 
After each up-sampling, we prune grid voxels by removing those with an SDF value higher/lower than threshold $\pm t
$, where $t\in \set{0.4, 0.9}$ enabling faster training and lower memory consumption; this is especially useful at the highest resolution grid. 
For 30 and 21 minutes running time budgets we prune with $t=0.9$, and for 15 and 8 minutes with $t=0.4$. For 30 minutes budget we perform 5 epochs at 64 resolution, 5 epochs at 128 and 3 at 256. For the rest of running time budgets we perform 2 epochs at each resolution, except for the 8 minutes budget, where at 256 resolution we perform only 1 epoch.
\rebb{Each epoch consists of 12800 iterations. At each training iteration the batch is composed by sampling random 10\% of the \emph{active} voxels (those which are left after pruning).}

For all the final experiments we set $\lambda_p=0.1$, $\lambda_n=10^{-5}$, $\lambda_v=1e-4$, $\lambda_c=1e-6$, $\epsilon=1e-2$. As for the optimizer, we use Adam \citep{kingma2014adam} with a constant learning rate of $0.001$, $\beta_1 = 0.9$ and $\beta_2 = 0.999$. All models are trained with a single NVIDIA Quadro GP-100 GPU.

\rebb{We did a grid search for all the hyper-parameters in the range of $10^{-6}$ to $10^{-1}$ with multiplicative steps of $10^{-1}$. We observed minimal performance difference. For all benchmark datasets we use the exact same hyper-parameters. More specifically, for the two proposed new losses -- Viscosity and Coarea -- we observe no performance change in the ranges $[5e-3, 5e-2]$ and $[5e-7, 5e-6]$, respectively, which allows for consistency across scenes with fixed hyper-parameters (see Fig. 7 and 8).}

We will publish the source code which reproduces the experimental results upon the paper acceptance.

\section{Training time}

In this section, we present qualitative and quantitative results of VisCo for different running time budgets. We experiment with the running time versus reconstruction quality trade-offs and show that short time training produces comparable reconstruction quality to longer time training.
In Tab. 1 we show quantitative results and in Fig. 1 qualitative results. Note that reducing the running time from 30 mins to 8 mins only marginally reduces reconstruction metrics, while qualitatively produces indistinguishable reconstruction results. The different running times versions were created mostly by reducing the number of epochs per resolution from $5$ down to $2$ (see more details in Sec.~\ref{s:implementation_details}). We strongly believe that further significant speedups are possible with a more efficient implementation. 

\rebb{
Below we report average time and memory footprint required for a single training iteration on NVIDIA Quadro GP100 GPU. Because of the pruning applied to the grid, we need to learn only a sparse set of the grid values (we call them \emph{active}).   
\begin{itemize}
\item $64^3$ resolution ($57\%$ of the grid values are active): 2.3 msec, 975MB VRAM
\item $128^3$ resolution ($31\%$ of the grid values are active): 8.6 msec, 1070MB VRAM
\item $256^3$ resolution ($30\%$ of the grid values are active): 25.8 msec, 1650MB VRAM
\end{itemize}
For neural networks (INRs) every point evaluation requires forward and backward in a network involving all network’s parameters in general, while for a grid we only require nearby grid function values (learnable parameters). Typical iteration times for NN (taken from DiGS) are:
\begin{itemize}
\item 66.5K params: 5.2-12.0 msec
\item 2.1M params: 17.5 msec
\end{itemize}

}
\begin{table}[t]\scriptsize	
    \centering
    \begin{tabular}{c|c|c|c|c|c|c||c}
        \multicolumn{2}{c|}{} & Anchor & Daratech & DC & Gargoyle & Lord Quas & Mean\\ \hline
        \multirow{4}{*}{30 min}
            & $d_C$ & \textbf{0.21} & \textbf{0.26} & \textbf{0.15} & \textbf{0.17} & \textbf{0.12} & \textbf{0.19} \\
            & $d_H$ & 3.00 & 4.06 & 2.22 & 4.40 & 1.06 & 2.95 \\
            & $d_C^\too$ & 0.15 & 0.14 & 0.09 & 0.11 & 0.07 & 0.11 \\
            & $d_H^\too$ & 1.07 & 1.76 & 2.76 & 0.95 & 0.64 & 1.44 \\ \hline
        \multirow{4}{*}{21 min}
            & $d_C$ &  \underline{0.27} & \underline{0.27} & \underline{0.16} & \textbf{0.17} & \textbf{0.12} & \underline{0.20} \\
            & $d_H$ & 5.60 & 4.06 & 2.13 & 4.33 & 0.99 & 3.42 \\
            & $d_C^\too$ & 0.14 & 0.14 & 0.09 & 0.11 & 0.07 & 0.12 \\
            & $d_H^\too$ & 1.17 & 1.77 & 2.77 & 0.93 & 0.64 & 1.45 \\ \hline
        \multirow{4}{*}{15 min}
            & $d_C$ & \underline{0.27} & \textbf{0.26} & \textbf{0.15} &  \textbf{0.17} & \textbf{0.12} & \underline{0.20} \\
            & $d_H$ & 5.68 & 4.20 & 2.23 & 4.45 & 1.10 & 3.53 \\
            & $d_C^\too$ & 0.14 & 0.14 & 0.09 & 0.11 & 0.07 & 0.11 \\
            & $d_H^\too$ & 1.10 & 1.77 & 2.80 & 1.04 & 0.66 & 1.47 \\ \hline
        \multirow{4}{*}{\textbf{8 min}}
            & $d_C$ & 0.28 & \textbf{0.26} & \textbf{0.15} &  \textbf{0.17} & \underline{0.13} & \underline{0.20} \\
            & $d_H$ & 5.69 & 4.15 & 2.23 & 4.45 & 1.14 & 3.53 \\
            & $d_C^\too$ & 0.15 & 0.13 & 0.09 & 0.11 & 0.07 & 0.11 \\
            & $d_H^\too$ & 1.15 & 1.78 & 2.78 & 0.98 & 0.68 & 1.48 \\ \hline
    \end{tabular} \vspace{10pt}
    \caption{Quantitative results of VisCo for different training time budgets on the surface reconstruction benchmark~\cite{williams2019deep}. Reducing the running time from 30 mins to 8 mins only marginally reduces reconstruction metrics.}
    \label{tab:ablations_time}\vspace{-10pt}
\end{table}

\begin{figure}[!t]%
  \begin{center}
  \includegraphics[width=\textwidth]{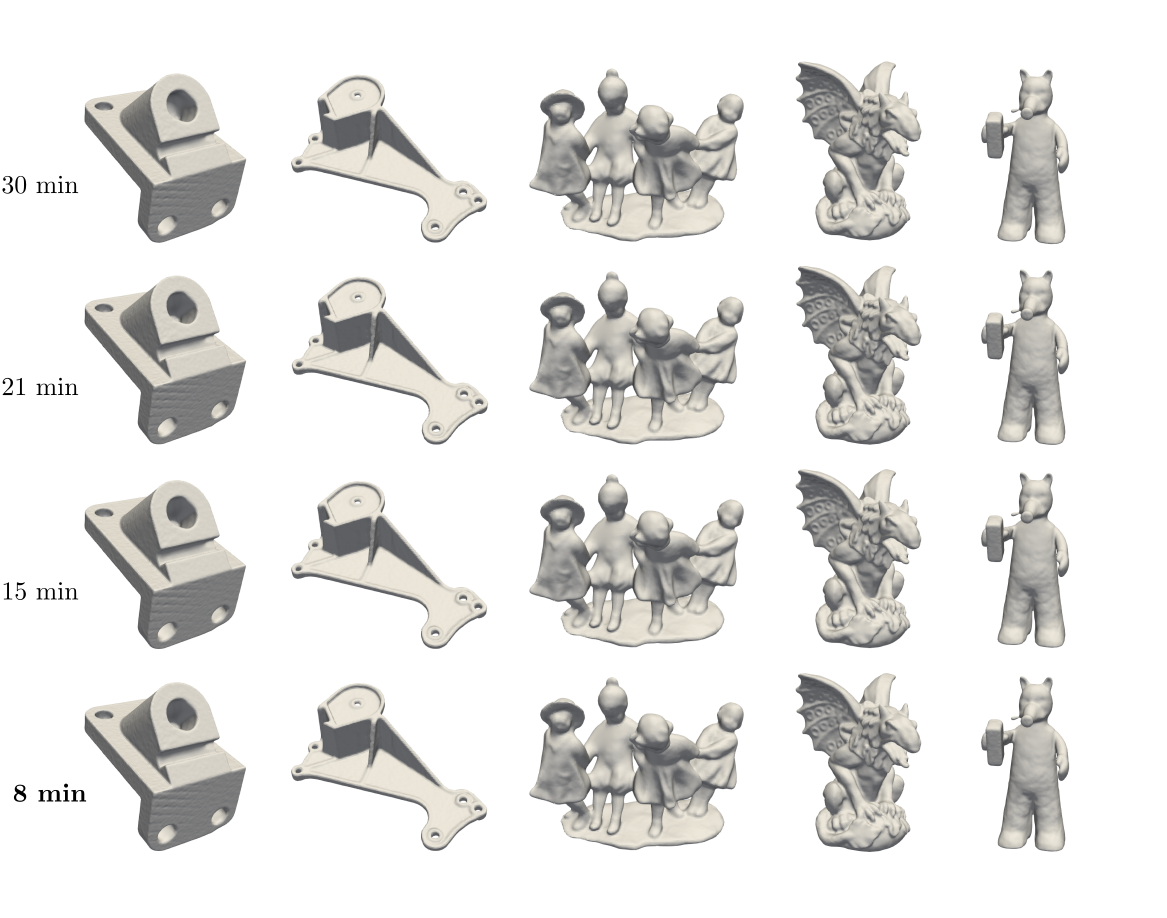}
  \end{center}\vspace{-20pt}
  \caption{Qualitative results of VisCo for different training time budgets on the surface reconstruction benchmark~\cite{williams2019deep}. Note that reduction of the training time does not result in inferior reconstruction. The models trained for 30 mins and 8 mins produce indistinguishable reconstruction results.}\label{fig:time}%
\end{figure}

\begin{figure}
  \begin{center}
  \includegraphics[width=0.9\textwidth]{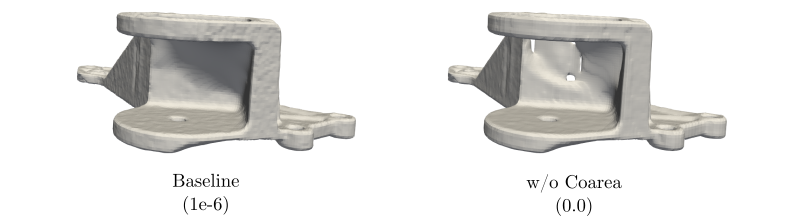}
  \end{center}
  \caption{\rebb{Visual comparison for Daratech between all loses vs. w/o Coarea. Reconstructed meshes from Tab. 3. Note small holes when removing the Coarea loss.}}\label{fig:coareavs}
\end{figure}

\rebb{
\section{Daratech Coarea Effect}\label{s:coareavs}
In this section we further study why in Tab. 3,  Daratesh seem to have better reconstruction w/o Coarea loss. Visual inspection reveals higher qualitative result for the mesh reconstructed with the Coarea loss although it has a higher quantiative error. We observe small holes when removing the Coarea loss, see Fig.~\ref{fig:coareavs}.
}

\end{document}